# Deep Machine Learning Based Egyptian Vehicle License Plate Recognition Systems

Mohamed Shehata
Systems & Computers Engineering Department, Faculty of Engineering Al-Azhar University, Egypt
+2 0122 113 6789
m-shehata@azhar.edu.eg

Mohamed Taha Abou-Kreisha
Mathematical & Computer Science Department, Faculty of Science, Al-Azhar University. Egypt.
+2 0100 252 8607
drkresha@gmail.com

Hany Elnashar
Project Development Unite, Faculty of Computers and Artificial Intelligent, Beni-Suef University
+2 0100 657 4960
hsoliman@fcis.bsu.edu.eg

## ABSTRACT

Automated Vehicle License Plate (VLP) detection and recognition have ended up being a significant research issue as of late. VLP localization and recognition are some of the most essential techniques for managing traffic using digital techniques. In this paper, four smart systems are developed to recognize Egyptian vehicles' license plates. Two systems are based on character recognition, which are (System1: Characters Recognition with Classical Machine Learning) and (System2: Characters Recognition with Deep Machine Learning). The other two systems are based on the whole plate recognition which are (System3: Whole License Plate Recognition with Classical Machine Learning) and (System4: Whole License Plate Recognition with Deep Machine Learning). We use object detection algorithms, and machine learning based object recognition algorithms. The performance of the developed systems has been tested on real images, and the experimental results demonstrate that the best detection accuracy rate for VLP is provided by using the deep learning method. Where the VLP detection accuracy rate is better than the classical system by 32%. However, the best detection accuracy rate for Vehicle License Plate Arabic Character (VLPAC) is provided by using the classical method. Where VLPAC detection accuracy rate is better than the deep learning-based system by 6%. Also, the results show that deep learning is better than the classical technique used in VLP recognition processes. Where the recognition accuracy rate is better than the classical system by 8%. Finally, the paper output recommends a robust VLP recognition system based on both statistical and deep machine learning.

## 1. INTRODUCTION

Vehicle License Plates Recognition (VLPR) system was invented in 1976 at the Police Scientific Development Branch in the UK but only in the late 90s became an important application domain of pattern recognition [1]. It plays a very important part in an intelligent transportation system (ITS) [2], and also plays a role in the decrease in the number of traffic violations, resulting in safer traffic flow. It's utilized to identify vehicles based only on their number plate. Since every car or truck has a special number plate, no further external cards, tags, or transmitters have to be recognized. It may be used in different road and security traffic management applications, like road electronic toll collection, average speed enforcement, parking control devices, detection of stolen cars, etc [3]. LPR system is an image processing as well as pattern recognition approach utilized to recognize vehicle license plate characters through a photo of a vehicle.

The paper is structured as follows: Section two delivers a briefing of relevant works. An overview of the complete proposed solutions is explained in section three. And also, the paper is concluded with Section five.

## 2. RELATED WORKS

In this section, a survey on automatic VLP detection and recognition techniques based on computer vision is presented. It reviews the implementation of new technologies in both the plate detection and character identification aspects of the process. All methods considered are from researches published in 2016 onwards.

Shaimaa Ahmed Elsaid, et al. [4], build a Real Time License Plates Recognition system that put on to Saudi Arabian VLPs. This method recognizes equally Arabic along with Indian numerals and also Latin alphabets and limited Arabic. The developed structure locates Saudi license plates in a shot picture no matter the time of day or license plate dimensions. It is composed of 5 primary stages; preprocessing, license plate localization, character segmentation, features extraction, along with character recognition utilizing Optical Character Recognition (OCR). To display the effectiveness of the proposed method, it was tested on 470 LP pictures taken in outdoor atmospheres such as numerous kinds of vehicles with various shadow, skewness and noise effects. The experimental results yield 96% segmentation accuracy as well as 94.7% recognition accuracy.

Rhen Anjerome Bedruz, et al. [5], proposed a vehicle plate optical character recognition technique using SIFT integrated fuzzy logic and image segmentation. Picture segmentation separates every single character in a plate region to get the features of any character obtained. Scale Invariant Feature Transform or even SIFT, on the other hand, allows for every feature of every character from the plate. Fuzzy reason analyzes the features from the SIFT algorithm that is suggested to identify the characters properly. This system used MATLAB to identify the functionality of the algorithm. the proposed algorithm helped remove plate character features along with knowing the characters in a certain picture. Results indicate the algorithm has an accuracy of 90.75%.

Ohnmar Khin, et al. [6], applied and also attempted to verify VLPR for Myanmar automobile quantity plates utilizing localization as well as recognition processes put on to license plates. Threshold based technique and bounding box techniques were used in various stages of the recognition process. This was tested for forty-eight letters as well as fifty-two numbers and also resulted in an accuracy rate of 90%.

Tejendra Panchal, et al. [7], addressed VLP limitation along with the integrated division technique. Various frameworks are provided for tag acknowledgment, each one bearing its specific purposes of blocks and intrigue. The serious stride of the VLPR framework is the actual repression of the number plate, Segmentation, Recognition. Harris corner computation is suggested to this method which ends up being effective in changing motion and enlightened lighting problems. Even though the exactness of VLP confinement is nourished forward to the segmentation organize. The Segmentation is enhanced by a method for associated segment investigation solidified with Pixel verify, Aspect proportion as well as Height of characters. The reenacted outcomes are coming out as well as the general accuracy was 93.84 %.

Jia Wang, et al. [8], developed a system for VLP detection and recognition called secondary positioning. It's used on the New Zealand license plate also it's dependent on Hue Saturation Value (HSV) color space. The first phase is founding the white light areas within the HSV color area, and after that, the precise location of the plate number is detected by learning the vertical edge of the plate number. For the knowing step, a template matching is applied. It contains a correction coefficient that's calculated between the templates as well as testing pictures. Accuracies, from plate number localization as well as the recognition, are above 75 % and 70 % respectively.

Hana M. Alyahya [9], designed a method that recognizes the plate number effectively. The bilinear interpolation algorithm was utilized to resize the license plate picture to get a fixed dimension of most license plates. Right after resizing the image, various preprocessing methods were applied to eliminate the needless parts. Lastly, an OCR which is grounded on the ANN classifier is utilized to recognize the characters. MATLAB device was used for developing and testing. Many experiments were conducted to compute accuracy and clearly show the effectiveness of the design system. The results showed that the system has 92 % accuracy and will recognize both Arabic and English letters and numbers.

Table 1 summariness the results of the mentioned related works.

**Table 1. Related Works VLP Recognition Systems Results**

| VLP Recognition Method | Accuracy |
|---|---|
| Hue Saturation Value [8] | 0.75 VLP |
| Threshold & bounding box [6] | 0.90 VLP |
| Fuzzy logic [5] | 0.91 VLP |
| ANN [9] | 0.92 VLP |
| Optical Character Recognition [4] | 0.94 VLP |
| Harris corner [7] | 0.94 VLP |

## 3. PROPOSED SYSTEMS

VLPR systems have an impressive spread both for their practical application and interest as a research issue. This section presents VLP detection and recognition systems based on computer vision techniques, which consists of foreground object extraction, VLP detection methods, VLP classification.

We identify Egyptian vehicle license plates instantly by four developed systems, making use of both deep and classical machine learning methods for license plate characters as well as a whole license plate. The input of these systems is a photo of a car that contains the license plate taken by a 2 MP (1920 × 1080) pixels resolution camera. Egyptian vehicle license plate styles are illustrated in Fig. 1.

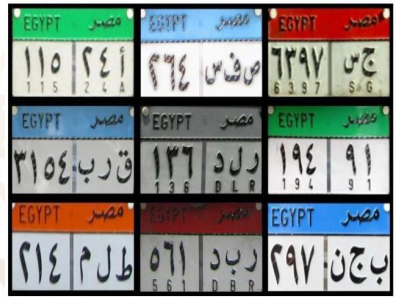

**Figure 1. Samples of Egyptian VLP design.**

Whereby the proposed systems assume the following:

- The plates have a rectangular shape.
- The width to height relationship of the license plate is known in advance.

- The plate has black characters on a bright white background
- The metal vertical bar separates the plate into two groups of characters.
- Items present in the plate center could be ignored, as they're unnecessary for the license plate recognition operation.
- The colored horizontal bars above the character's location might be ignored

These systems will be discussed in details in the following lines. Fig. 2, shows the model of the overall system.

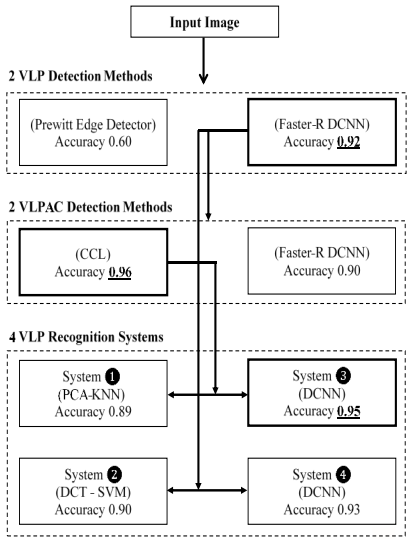

**Figure 2. Overall Systems Model.**

Two methods are developed for VLP detection and two methods are developed for VLPAC detection. The four methods are based on both statistical and deep learning techniques.

### 3.1 VLP Detection Methods
The primary phase begins with VLP extraction, and that is the primary key stage in VLPR systems. The goal of this particular stage is producing a VLP region as shown in Fig. 3.

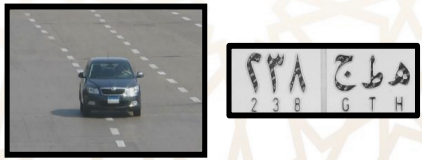

**Figure 3. VLP Detection Input & Output.**

We developed two methods for VLP detection. One of them is based on statistical techniques and the other one is based on deep learning techniques.

#### 3.1.1 VLP detection based on a statistical technique
To get the rectangular shape which has the identical attributes as the license plate; the algorithm initially converts the input color image to greyscale, then a Prewitt edge detector [10] is applied that returns a binary image of the same dimensions as the input image, then a dilation process is performed for those horizontal and vertical lines detected to avoid broken lines. And then, the algorithm discards minor lines that are not likely to be a component of the plate borders using a label connected components technique. Afterward, the algorithm checks for just about any horizontal lines linking two vertical lines which represent the vertical boundaries of the number plate, taking into account the license plate height to width ratio.

Out of the previous section, we can compute the VLP height and width, additionally, we can identify the upper left corner of the VLP, although it still not enough for the extraction process, as parts of some characters might be lost as an outcome of non-alignment of the license plate with the horizontal axis as shown in Fig. 4.

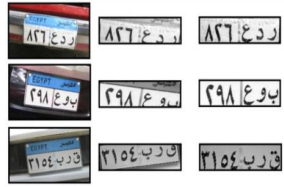

**Figure 4. Examples of License Plate Alignments.**

This might happen because of the camera as well as angle, defects in the fixation of the plate, or deformities of the plate as an outcome of traffic accidents.

After the plate is located, the alignment algorithm starts the preparation for the following stage. The whole idea depends upon the detection of the horizontal line within the license plate, demonstrated in Fig. 5.

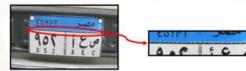

**Figure 5. The Horizontal Line Inside the License Plate.**

To detect that line we followed the next steps. The Horizontal Prewitt edge detected lines, which represent the top, bottom, along with tiny lines are discarded, as shown in Fig. 6.

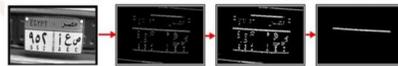

**Figure 6. Horizontal Prewitt Edge Detection.**

Next a novel rotation algorithm uses to compute the necessary rotation angle for the successful alignment of the horizontal line within the license plate with the horizontal axis. The rotation algorithm rotates the input image around its center stepwise by angles from 10 to -10, in steps of 0.5. At each step, the algorithm calculates the height difference between the upper left corner and the upper right corner of VLP, as shown in Fig. 7.

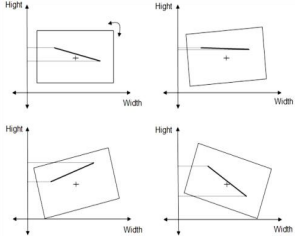

**Figure 7. License Plate Alignments Process.**

The minimum difference recorded, represents the required rotation angle; finally, we can extract the VLP region utilizing the resulting angle.

A number of 1726 VLPs have been tested. Table 2 shows the confusion matrix and the results of that method.

**Table 2. VLP Detection Based on Edge Detector Results**

| FN | FP | TP | Precision | Recall | Accuracy |
|---|---|---|---|---|---|
| 519 | 274 | 1207 | 0.82 | 0.70 | 0.60 |

*3.1.2 VLP detection based on deep learning technique*

Faster region deep convolutional neural network (Faster R-DCNN) detector [11] is used to find candidate bounding boxes. To locate VLP.

Faster R-DCNN has two networks, Region Proposal Network (RPN) for generating region proposals and a Deep Convolutional Neural Network (DCNN) using these proposals to detect objects. RPN ranks region boxes called anchors and proposes the ones most likely containing objects.

Anchors play an important role in Faster R-DCNN. An anchor is a box. In the default configuration of Faster R-DCNN, there are 9 anchors at a position of an image. Fig. 8, shows 9 anchors shapes.

The first step of training a classifier is to make a training dataset. The training data is that the anchors we get from the above process and therefore the ground-truth boxes. The problem we'd like to unravel here is how we use the ground-truth boxes to label the anchors. The basic idea here is that we would like to label the anchors having the upper overlaps with ground-truth boxes as foreground, those with lower overlaps as background. It needs some tweaks and compromise to separate the foreground and background. Now we have labels for the anchors.

Let's say the 1920X1080 image shrinks 32 times to a 60x34 feature map after applying DCNNs. Every position in the feature map has 6 anchors, and every anchor has two possible labels (background, foreground). If we make the depth of the feature map 12 (6 anchors x 2 labels), we will make every anchor have a vector with two values representing the foreground and background. If we feed these vectors into a SoftMax logistic regression activation function, it will predict the labels. Now the training data is complete with features and labels.

In Faster R-DCNN, receptive fields of various anchors often overlap one another, as you'll from the above graph. It leaves the RPN to be position-aware. After RPN, we get proposed regions with different sizes. a special sized region means different sized DCNN feature maps. It's tough to form an efficient structure to figure on features of various sizes. Region of Interest (ROI) Pooling can simplify the matter by reducing the feature maps to an equivalent size. Unlike Max-Pooling which features a fixed size, ROI Pooling splits the input feature map into a hard and fast number of (k) of roughly equal regions, then apply Max-Pooling on every region. Therefore, the output of ROI Pooling is usually (k) no matter the dimensions of the input.

At last, an R-DCNN takes input from both the DCNN feature network and RPN and generates the ultimate class and bounding box. R-DCNN consists of 4 fully connected, two of them stacked common layers shared by a classification layer and a bounding box regression layer, to assist it to classify only the within of the bounding boxes, the features are cropped consistent with the bounding boxes, to get the detected bounding boxes for the VLP. Fig. 9 [11], summarizes the Faster R-DCNN process.

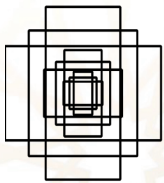

**Figure 8. Faster R-DCNN Anchor Boxes.**

The input image is (1920, 1080) and we choose one position at every stride of 32, there will be 2040 (60x34) positions. This leads to 18360 (2040 x 9) boxes to consider. The sheer size is hardly smaller than the combination of the sliding window and pyramid. Our designed network does not need to consider the square boxes. For that, we reduce the number of anchors to be 6 to increase the speed of the system.

The output of RPN is a bunch of boxes that will be examined by a classifier and regressor to eventually check the occurrence of vehicles, VLP, or VLPC. As the RPN predicts the possibility of an anchor being background or foreground, and refine the anchor.

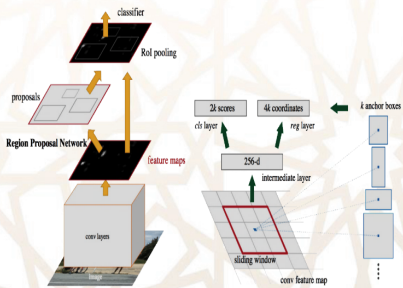

**Figure 9. Faster R-DCNN Process.**

We used 5000 Egyptian VLPs images taken from different angles of view with different types and sizes. We did some augmented effects on the VLPs images as we rotate them with different angles and deform them a little bit, which leads to train a Faster R-DCNN object detector by 100000 original and augmented

VLPs images, for detecting a VLP. Table 3 shows the confusion matrix and the results of that method.

**Table 3. VLP Detection Based on Faster R-DCNN Results**

| FN | FP | TP | Precision | Recall | Accuracy |
|---|---|---|---|---|---|
| 186 | 162 | 3859 | 0.96 | 0.95 | 0.92 |

## 3.2 VLPAC Detection Methods

We developed two robust methods based on statistical and deep learning techniques, that aim to extract the Arabic letters and Hindi digits within VLP, where the models' input is the VLP image which the previous system considered. And the system output should be the boundary boxes of the VLPAC. As shown in Fig. 10.

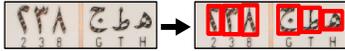

**Figure 10. VLPAC Detection Input & Output.**

### 3.2.1 VLPAC detection based on a statistical technique

This particular stage input would be the VLP that the previous method considered, and tries to find Arabic letters as well as Hindi digits inside these regions. To speed up the procedure we resized the input image to a lower resolution sample.

Before any processing or segmentation on a VLP image may be done, it's really helpful to convert the image from a grey scale or color to a white and black bitmap. This's a straightforward procedure that entails just scanning the image and changing each pixel value to either white or black based on if they're above or below a certain threshold. Otsu's method [12] was used to compute this particular threshold amount value.

When there are any ripples on any character such as a mark over character, the character will appear as artificially segmented. As a workaround, a dilation procedure follows in both axes. Utilizing the Connected Components Label (CCL) method [13] to count objects within separate character regions. Small objects like ripples as well as English characters in the bottom part of the characters region could subsequently be recognized utilizing ratio checks between the height and the width of the VLP. Small objects are then discarded to boost the effectiveness of the recognition process. To restore these lost areas after obtaining the required characters, we gradually expanded the capture area around every recognized character in all directions as shown in Fig. 11.

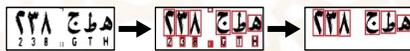

**Figure 11. VLPAC Detection.**

Then each character is going to be cropped into a block that contains no extra white spaces in all of the four sides of the characters.

A number of 7434 VLPACs have been tested. Table 4 shows the confusion matrix and the results of that method.

**Table 4. VLPAC Detection Based on CCL Results**

| FN | FP | TP | Precision | Recall | Accuracy |
|---|---|---|---|---|---|
| 196 | 124 | 7238 | 0.98 | 0.97 | 0.96 |

### 3.2.2 VLPAC detection based on deep learning

Faster R-DCNN detector which is mentioned in Section 3.1.2 is used to find candidate bounding boxes. To locate VLPAC. But here the designed network does not need to consider the horizontal anchor boxes and need the square boxes. For that, we reduce the number of anchors to be 6 to increase the speed of the system.

We used 13500 Egyptian VLPACs images taken from different 5000 VLP with different types and sizes. We did some augmented effects on the VLPACs images as we rotate them with different angles. Plus, we made a little deformation on the characters' images, which leads to train a Faster R-DCNN object detector by 270000 original and augmented VLPACs images, for detecting a VLPACs. A number of new 24714 VLPACs belongs to 4045 VLP have been tested. Table 5 shows the confusion matrix and the results of that method.

**Table 5. VLPAC Detection Based on Faster R-DCNN Results**

| FN | FP | TP | Precision | Recall | Accuracy |
|---|---|---|---|---|---|
| 1805 | 784 | 22609 | 0.96 | 0.92 | 0.90 |

## 3.3 System 1: Characters Recognition with Classical Machine Learning

This system depends on a sequential, multistage image processing method for the License Plate Extraction as well as Characters Segmentation. as shown in Fig. 12.

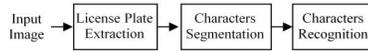

**Figure 12. The workflow of VLPR System Based on Character Recognition.**

This system input is the VLPACs that were detected before. The system output is the characters' names as shown in Fig. 13.

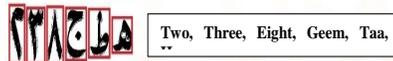

Two, Three, Eight, Geem, Taa,

**Figure 13. VLPR System Based on Character Recognition Input & Output.**

As feature extraction is the most remarkable part of a recognition system that has a significant impact on the recognition performance, we use Principal Component Analysis (PCA) for feature extraction [14].

The PCA may be a popular dimensionality reduction technique, PCA has been used widely in computer vision applications. it's a ubiquitous statistical procedure for unsupervised feature extraction, data analysis, and compression. it's a really extensive literature, a summary of which may be found in several monographs. Normally in PCA, the target function τ for choosing new directions is defined by

$$\tau(v) = \frac{v^T C v}{v^T v} \quad \text{Eq. 1}$$

Where

$$C = \mu\{(x - \mu\{x\})(x - \mu\{x\})^T\} \quad \text{Eq. 2}$$

Is the sample covariance matrix. Here $\mu$ denotes the mean value. The following proposition asserts a well-known property of the first equation, that defines $\tau(v)$ as the variance of the centralized sample vectors

$$x_1 - \mu\{x\}, \ldots, x_k - \mu\{x\}$$

Projected onto vector $\frac{v}{\|v\|}$.

Hence, this method prefers directions that have an outsized variance. Fig. 14, shows the effect of PCA on 2D data sets, the synthetic data sets on the right-hand side were transformed using PCA and therefore the results are shown on the left-hand side. Besides the info sets themselves, we also plotted their distribution along the axes.

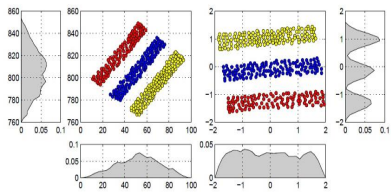

**Figure 14. PCA Effect on 2D Data Sets.**

A dataset of characters used to train the PCA feature extraction function. A sample of that data set is shown in Fig. 15.

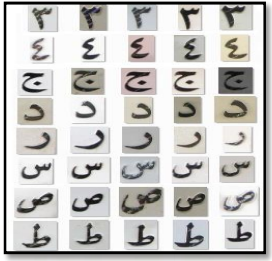

**Figure 15. Samples of Characters Trained by PCA.**

Then the features extracted from the input character images are compared to the features extracted from characters images in a training database, and the best similarity is considered. To measure the similarity, a minimum distance classifier K-NN Euclidean distance [15] is used. K- NN are often computed through computing the space between the test image and therefore the training images, a matrix during which each image is represented as a vector. then find the minimum distance between the test image and therefore the training images. the worth of the minimum distance decides the group of the test image. On the opposite hand, the minimum distance classifier is often computed by calculating the minimum distance vector. The classifier is based on many distance measures. After arranging the feature vectors of the training images into a matrix $X$ of $N$ column vectors $X_1, X_2, \ldots, X_n$, the distance D between a given test feature vector $X_{tst}$ and each of the training feature vectors $X_i$ is calculated using various distance metrics. Euclidean Distance is one among the foremost commonly used distance measures. It is the square root of the sum of the squared distances of two vector values $(X_{tst}, X_i)$. The Euclidean distance are often seen because the shortest distance between two points. it's defined as:

$$D_e = \sqrt{(X_{tst} - X_i)(X_{tst} - X_i)^T} \qquad \text{Eq. 3}$$

Another metric is the standardized Euclidean distance. It is defined as [118]:

$$D_s = \sqrt{(X_{tst} - X_i)d^{-1}(X_{tst} - X_i)^T} \qquad \text{Eq. 4}$$

Where $d^{-1}$ is the diagonal matrix with diagonal elements given by $v_j^2$, and $v_j^2$ is the variance of the $j_{th}$ the element of $X$. The advantage of the Standardized Euclidean metric is that it takes into account the variance mentioned above. This distance is widely used in 2D metrics but can also be highly useful in n-dimensional vectors such as feature vectors due to their low computational costs.

10 images for 9 digits and 17 alphabets with a total number of 260 characters have been used for the training process, and a new 300 VLP contains 1873 characters have been tested. The system precision is 0.85, recall is 0.84, and accuracy is 0.97 for character recognition, which leads to a VLP recognition accuracy rate of 0.89.

### 3.4 System 2: Characters Recognition with Deep Learning

This system utilized the identical workflow of (System 1) as shown in Fig. 12, besides it's used a full DCNN [16] instead of the DCT method that is needed for feature extraction as well as a KNN classifier.

In the field of deep learning, DCNN has performance that is excellent for visual recognition. In this system, DCNN AlexNet. The DCNN basic architecture is shown in Fig. 16 [17].

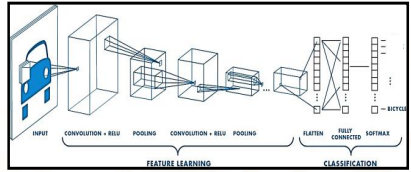

**Figure 16. DCNN Basic Architecture.**

The DCNN first layer defined the sort also because the size of the input character image; next, we defined the center layers of the system. the center layers are comprised of repetitive blocks of convolution layers, which may be liable for executing the image manipulation procedures with the convolution functions to control also as extract the character image features. Every convolution layer is followed by a cross channel normalization layer, a Rectified Linear Measure (ReLU), alongside a pooling layer to rework the convolution process results and also create the DCNN invariant to image translation and illumination. These layers make the core building blocks of DCNN. The last layers are the classification layers and have 26 classes which represent all characters utilized in the Egyptian car place. For the classification process, the input size must be the standard size of the training images. Two fully connected neural network layers are used. These layers combine all the features learned by the previous layers across the image to acknowledge the larger patterns.

The proposed network utilized to train a database consisted of 270000 photos for license plate characters which were used before in section 3.2.2. The New 300 VLP contains 1873 characters that have been tested. The system precision is 0.95, recall is 0.95, and accuracy is 0.99 for character recognition, which leads to a VLP recognition accuracy rate of 95%.

### 3.5 System 3: Whole License Plate Recognition with Classical Machine Learning

In this system, we excluded the character segmentation stage that had been utilized before in the previous two recognition systems, as shown in Fig. 17. Right after, extracting the VLP, as we immediately recognize it, using a feature extraction algorithm, along with classical machine learning based classification algorithms.

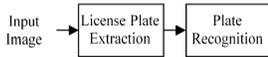

Figure 17. LPR system based on whole license plate recognition.

Whereas we followed the same image processing steps, that are pointed out in the VLP detection stage until the VLP extraction takes place. as we do not need a process for character segmentation. Rather, we found features in the whole VLP image.

Where the system input is the VLP image and the system output is the recognized VLP as shown in Fig. 18.

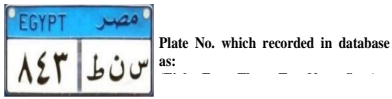

Figure 18. VLPR System Input & Output.

After detecting the VLP as illustrated before, we directly extract it and start to recognize it, using a feature extraction algorithm, and classical machine learning based classification algorithms.

We don't need the character segmentation process. Instead, we found features in the whole VLP image $f_i(x, y)$. Each image is $N_x \times N_y$. DCT features extraction technique is used, by dividing the input image into equally sized non-overlapped blocks 8x8 pixels. Each block is transformed from the spatial domain to the frequency domain and represented by its Discrete Cosine Transform (DCT) coefficients [18].

$$B_{pq} = \alpha p \alpha q \sum_{x=0}^{N_x-1} \sum_{y=0}^{N_y-1} f(x,y) \cos\frac{\pi(2x+1)p}{2N_x} \cos\frac{\pi(2y+1)q}{2N_y} \quad \text{Eq. 5}$$

At $0 \leq p \leq N_x - 1$ and $0 \leq q \leq N_y - 1$

where f(x,y) Input image, $N_x$ Input image rows number, and $N_y$ Input image columns number.

$$\alpha p = \begin{cases} \sqrt{\frac{1}{N_x}} & p = 0 \\ \sqrt{\frac{2}{N_x}} & 1 \leq p \leq N_x - 1 \end{cases} \quad \text{Eq. 6}$$

$$\alpha q = \begin{cases} \sqrt{\frac{1}{N_y}} & q = 0 \\ \sqrt{\frac{2}{N_y}} & 1 \leq q \leq N_y - 1 \end{cases} \quad \text{Eq. 7}$$

From the remaining DCT coefficients, the ones containing the highest information are extracted via zigzag scan [18], with the most visually significant information being concentrated in the first few DCT coefficients ordered in a zigzag pattern from the top left corner as shown in Fig. 19.

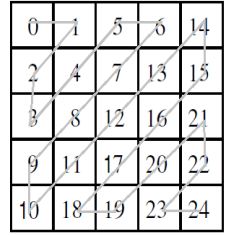

Figure 19. Zigzag Sequence.

Information concentration in the top left corner is due to the correlation between image and DCT properties. For images, the most visually significant frequencies are of low-mid range, with higher frequencies giving finer details of the image. For the 2D DCT, coefficients correspond to sinusoids of frequencies increasing from left to right and top to bottom. Therefore, those in the upper corner are the coefficients associated with the lowest frequencies. The first coefficient, known as the DCT coefficient, represents the average intensity of the block and is the most affected by variation in illumination. DCT features have been used in a holistic appearance-based sense, or local appearance-based sense that ignores spatial information during the classification step [19].

The DCT coefficients obtained from each block are concatenated to construct the feature vector that is used by the classifier. The previous process is performed on all the training images. To identify unknown images the following steps will be applied.

After that, the features extracted from the input VLP images are compared to the features extracted from a set of VLP in a training database, and the best similarity is considered by using the SVM classifier [20].

SVM is an unsupervised approach based on statistical learning theory. It estimates the optimal boundary in the feature space by combining a maximal margin strategy with a kernel method; this process is called a kernel machine. The machine is trained according to the structural risk minimization criterion. The decision boundaries are directly derived from the training data set by learning. The SVM maps the inputs into a high-dimensional feature space through a selected kernel function. Then, it constructs an optimal separating hyper-plane in the feature space. The dimensionality of the feature space is determined by the number of support vectors extracted from the training data shown in Fig. 20.

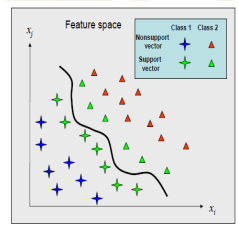

Figure 20. Optimal Boundary Searched by SVM.

The SVM can locate all the support vectors, which exclusively determine the decision boundaries. To estimate the misclassification rate (risk), the so-called leave-one-out procedure is used. It removes one of $N_i$ training samples, perform training using the remaining training samples and tests the removed sample with the newly derived hyper-plane. It repeats this process for all of the samples, and the total number of errors becomes the estimation of the risk.

A number of 500 VLPs images from various angles of views, for 100 different vehicles, used for training. And new 100 VLPs have been tested. The system precision is 0.85, recall is 0.76, and accuracy is 0.90.

### 3.6 System 4: Whole License Plate Recognition with Deep Learning

This system utilized the same workflow illustrated in (System3) as shown in Fig. 17, where, we excluded the character segmentation stage that had been utilized before. Right after, extracting the VLP, as we immediately recognize it, using a DCNN which mentioned in (System 2) for feature extraction, and two fully connected neural network layers as a classifier, to recognize the license plate.

We used 5000 VLP images of different types and sizes. We did some augmented effects on the VLP images as we rotate them with different angles. Plus, we made a little deformation on the VLP images, which leads to train a DCNN by 100000 original and augmented VLPs images, divided into 5000 classes. A number of new 1000 VLP have been tested. The system precision is 0.91, recall is 0.89, and accuracy is 0.93.

### 4. CONCLUSIONS

We've talked about the concepts of VLPR in the Egyptian environment. We proposed two methods for VLP detection and two methods for VLPAC detection, based on statistical and deep learning methods. Then we created four comprehensive smart systems to identify as well as recognize Egyptian license plates. Finally, we conclude that the most accurate VLPR proposed system which has the highest accuracy. It should use Faster R-DNN deep learning method for VLP extraction followed by CCL statistical method for VLPAC detection. At last, a DCNN is utilized for character recognition prosses.